\documentclass[11pt,a4paper]{article}
\usepackage[hyperref]{acl2020}
\usepackage{times}
\usepackage{latexsym}

\usepackage[ruled]{algorithm2e}

\usepackage{algorithmic}

\usepackage{pgfplots}
\usepackage{microtype}

\aclfinalcopy 

\title{DARE: Data Augmented Relation Extraction with GPT-2}

\author{Yannis Papanikolaou and Andrea Pierleoni \\
  Healx, Cambridge, UK \\
  \texttt{\{yannis.papanikolaou, andrea.pierleoni\}@healx.io} 
\\}

\date{}

\begin{document}
\maketitle
\begin{abstract}
Real-world Relation Extraction (RE) tasks are challenging to deal with, either due to limited training data or class imbalance issues. In this work, we present \textit{Data Augmented Relation Extraction} (DARE), a simple method to augment training data by properly fine-tuning GPT-2 to generate examples for specific relation types. The generated training data is then used in combination with the gold dataset to train a BERT-based RE classifier. In a series of experiments we show the advantages of our method, which leads in improvements of up to 11 F1 score points against a strong baseline. Also, DARE achieves new state of the art in three widely used biomedical RE datasets surpassing the previous best results by 4.7 F1 points on average.
\end{abstract}

\section{Introduction}

Relation Extraction (RE) is the task of identifying semantic relations from text, for given entity mentions within it. This task, along with Named Entity Recognition, has recently become increasingly important due to the advent of knowledge graphs and their applications. In this work, we focus on supervised RE \citep{zeng2014relation, lin2016neural, wu2017adversarial, verga2018simultaneously}, where relation types come from a set of predefined categories, as opposed to Open Information Extraction approaches that represent relations among entities using their surface forms \citep{banko2007open, fader2011identifying}.

RE is inherently linked to Natural Language Understanding in the sense that a successful RE model should manage to adequately capture language structure and meaning. So, almost inevitably, the latest advances in language modelling with Transformer-based architectures \citep{radford2018improving, devlin2018bert, radford2018language} have been quickly employed to also deal with RE tasks \citep{soares2019matching, lin2019bert, shi2019simple, papanikolaou2019deep}.

These recent works have mainly leveraged the discriminative power of BERT-based models to improve upon the state of the art (SOTA). In this work we take a step further and try to assess whether the text generating capabilities of another language model, GPT-2 \citep{radford2018language}, can be applied to augment training data and successfully deal with class imbalance and small-sized training sets.


\begin{table*}
\centering
\begin{tabular}{cc}
\hline 
Dataset(relation type)&Generated sentences \\ 
\hline
\hline
CDR(Induce)&DISEASE was the most common adverse reaction ( 21 \% ) reported for DRUG,\\
&and occurred in approximately 50 \% of patients .\\
DDI2013(Effect)&DRUGA may enhance the effects of alcohol, barbiturates, DRUGB, and other \\
&cns depressants. \\
DDI2013(Advise)&caution should be observed when DRUGA and DRUGB are \\
&coadministered.\\
DDI2013(Mechanism)&co-administration of DRUGA decreased the oral bioavailability (48\%) of\\
&DRUGB, a substrate for cyp2d6.\\
ChemProt(Activate)&DRUG enhances PROTEIN sensitivity via activation of the pi3k / akt\\
&signaling pathway. \\
ChemProt(Inhibit)&DRUG, a novel orally bioavailable xanthine PROTEIN inhibitor,\\
ChemProt(Product)&the enzyme PROTEIN catalyzes the two-electron reduction of DRUG to\\
&produce acetyl groups.\\
\hline 
\end{tabular}
\caption{\label{tbl:examples} Examples of generated sentences with fine-tuned GPT-2 models. Each model is fine-tuned on examples from the specific relation type.}
\end{table*}

Specifically, given a RE task we fine-tune one pretrained GPT-2 model per relation type and then use the resulting fine-tuned models to generate new training samples. We then combine the generated data with the gold dataset and fine-tune a pretrained BERT model \citep{devlin2018bert} on the resulting dataset to perform RE. 

We conduct extensive experiments, studying different configurations for our approach and compare DARE against two strong baselines and the SOTA on three well established biomedical RE benchmark datasets. The results show that our approach yields significant improvements against the rest of the approaches. To the best of our knowledge, this is the first work augmenting training data with GPT-2 for RE. In Table \ref{tbl:examples} we show some generated examples with GPT-2 models fine-tuned on the datasets that are used in the experiments (refer to Section \ref{sec:datasets}).

In the following, we provide a brief overview of related works in Section \ref{sec:related_work}, we then describe our approach in Section \ref{sec:method}, followed by our experimental results (Section \ref{sec:evaluation}) and the conclusions (Section \ref{sec:conclusions}).

\section{Related Work}
\label{sec:related_work}

Relation Extraction is usually modelled as a text classification task. Therefore most methods to deal with class imbalance or limited data in RE follow the respective methods from text classification. In the following, we describe the different approaches that have been followed in the literature. 

One approach is to deal with imbalance at the classifier level, by penalizing misclassification errors differently for each class, depending on the class frequency \citep{lewis2004rcv1, zhou2005training} or by explicitly adjusting prior class probabilities \citep{lawrence1998neural}.

Another popular approach relies on either undersampling the majority class(es) or oversampling the minority one(s), transforming the training data with the aim of balancing it. One of the simplest approaches, random majority undersampling, simply removes a random portion of examples from majority classes so that per class training examples are roughly equal \citep{japkowicz2002class}. An improved version of the previous method, balanced bagging \cite{hido2009roughly}, employs an ensemble of classifiers that have been trained with random majority undersampling. 

Oversampling approaches for textual data have been somehow limited as opposed to those for image data \citep{wong2016understanding, fawzi2016adaptive, wang2017effectiveness, frid2018gan}, since text semantics depend inherently on the exact order or structure of word tokens. 

A simple approach is to replace words or phrases with their synonyms \citep{zhang2015character}. \citet{chen2011exploiting} employed topic models to generate additional training examples by sampling from the topic-word and document-topic distributions. \citet{ratner2016data} proposed a data augmentation framework that employs transformation operations provided by domain experts, such as a word swap, to learn a sequence generation model. \citet{kafle2017data} used both a template-based method and an LSTM-based approach to generate new samples for visual question answering.

A similar method to our approach was proposed by \citet{sun2019lamal} who presented a framework to successfully deal with catastrophic forgetting in language lifelong learning (LLL). Specifically and given a set of tasks in the framework of LLL, they fine-tune GPT-2 to simultaneously learn to solve a task while generating training samples for it. When dealing with a new task, the model is trained on the generated training samples from previous tasks alongside the data of the new task, therefore avoiding catastrophic forgetting.

Our work falls into the oversampling techniques for text, but our focus is RE. Importantly, we do not need any domain expertise, templates, synonym thesaurus or to train a model from scratch, which makes our approach easily adaptable to any domain, with relatively low requirements in resources.

\section{Methods}
\label{sec:method}
In this section we present briefly the GPT-2 model and before giving a detailed introduction to our approach.
\subsection{GPT-2}
GPT-2 \citep{radford2018language} is a successor of the GPT language model \citep{radford2018improving}. Both models are deep neural network architectures using the Transformer \citep{vaswani2017attention}, pre-trained on vast amounts of textual data. Both models are pre-trained with a standard language modelling objective, which is to predict the next word token given $k$ previously seen word tokens. This is achieved by maximizing the following likelihood:
\begin{equation}
    L(U) = \sum_i log P(u_i|u_{i-1},...,u_{i-k};\Theta)
\end{equation}
where $\Theta$ are the neural network parameters. The authors have gradually provided publicly four different flavours of GPT-2, with 124M, 355M, 774M and 1558M parameters respectively. In our experiments we use the second largest model (774M), since it seems to represent a good compromise between accuracy and hardware requirements\footnote{https://openai.com/blog/gpt-2-1-5b-release/}.

\subsection{Data Augmented Relation Extraction}
Let $D = [s_0,...s_d]$ be a RE dataset containing $d$ sequences. Furthermore, we assume that each sequence ${s} = [w_0,...w_n]$ will be a sequence of $n$ word tokens and that ${e_1} = [w_{e1i},...w_{e1j}]$ and ${e_2} = [w_{e2k},...w_{e2l}]$ will represent a pair of entity mentions in $s$. Furthermore, let $L = [l_1, ..., l_c]$ be a set of $c$ relation types. Then, RE is the task of learning a function that maps each triple $(s_i, e1, e2)$ to $L$, i.e.,
\begin{equation}
    h = f_{\Theta}(s_i, e1, e2), h \in L
\end{equation}
where $\Theta$ are the parameters of the model.

In this work we employ a RE classifier based on a pretrained BERT language model. This classifier follows the same principle followed by \citet{devlin2018bert}, using a special token (CLS) for classification. The only modification is that we mask entity mentions with generic entity types, i.e., \textit{\$ENTITY\_A\$} or \textit{\$ENTITY\_B\$}. It should be noted that the method that we introduce here is not classifier specific, so any other classifier can be used instead.

To generate new training data, we split the $D$ dataset into $c$ subsets where each $D_c$ subset contains only examples from relation type $c$. Subsequently, we fine-tune GPT-2 on each $D_c$ for five epochs and then prompt each resulting fine-tuned model to generate new sentences, filtering out sentences that do not contain the special entity masks or that are too small (less than 8 tokens). The generated sequences are combined for all relation types into a dataset $Dsynth$.

Subsequently, we build an ensemble of RE classifiers, each of them being fine-tuned on a subset of $Dsynth$ and the whole $D$, such that the per-relation type generated instances are equal to the number of gold instances for that relation, multiplied by $ratio$, i.e., $|Dsynth'_c| = |D_c|*r$. In our experiments we have set $r=1.0$ (refer to Section \ref{sec:ratio_study} for a short study of its influence). Algorithm \ref{alg:dare} illustrates our method.
\begin{algorithm}[!ht]
\SetAlgoNoLine
\KwIn{$D$, $L$}
\BlankLine
\For{each relation type $c \in L$} 
{	
	$D_c = \{s \mid s\in D , rel\_type(s) = c\}$\;
	fine-tune GPT-2 on $D_c$\;
	generate $Dsynth_c$ with GPT-2\;
}
$Dsynth = Dsynth_{1}\cup ... \cup Dsynth_{c}$\;

\For{each classifier in ensemble} 
{	
    $Dsynth' \sim Dsynth$ s.t. $|Dsynth'_c| = r*|D_c|$ and $Dsynth' = Dsynth'_{1}\cup ... \cup Dsynth'_{c}$\;
	train RE classifier on $D\cup Dsynth'$\;
}

predict on $Dtest$ with majority voting over the ensemble\;

\BlankLine
\caption{DARE}
\label{alg:dare}
\end{algorithm}

We would like to note that in early experiments, we also experimented with fine-tuning over the whole $D$, by adding a special token to the beginning of each sentence that encoded the relation type, e.g., $<$0$>$: or $<$1$>$:. Then during generation, we would prompt the model with the different special tokens and let it generate a training instance from the respective relation type. However, this approach did not prove effective leading to worse results than just using gold data, primarily because frequent classes "influenced" more GPT-2 and the model was generating many incorrectly labeled samples.

\section{Experimental Evaluation}
\label{sec:evaluation}

In this section we present the empirical evaluation of our method. We first describe the experimental setup, the datasets used, the baselines against which we evaluate DARE and subsequently present the experiments and report the relevant results. 

\label{sec:datasets}
\begin{table*}
\centering
\begin{tabular}{ccccc}
\hline 
Dataset&$|L|$&Training&Development&Test \\ 
\hline
\hline
CDR&1&3,597(1,453)&3,876&3,806 \\ 
DDI2013&4&22,501(153 658 1,083 1,353)&4,401&5,689 \\ 
ChemProt&5&14,266(173 229 726 754 2,221)&8,937&12,132 \\ 
\hline 
\end{tabular}
\caption{\label{tbl:datasets} Statistics for the datasets used in the experiments. For the training data we provide in parentheses the number of positives across each class. We do not include in $|L|$ the \textit{null} class which signifies a non-existing relation.}
\end{table*}

\subsection{Setup}
\label{sec:setup}
In all experiments we used the second-largest GPT-2 model (774M parameters). All experiments were carried out on a machine equipped with a GPU V100-16GB. For the implementation, we have used HuggingFace's Transformers library \citep{Wolf2019HuggingFacesTS}.

To fine-tune GPT-2 we employed Adam as the optimizer, a sequence length of 128, a batch size of 4 with gradient accumulation over 2 batches (being equivalent to a batch size of 8) and a learning rate of $3e-5$. In all datasets and for all relation types we fine-tuned for 5 epochs. For generation we used a temperature of $1.0$, fixed the top-k parameter to $5$ and generated sequences of up to 100 word tokens. An extensive search for the above optimal hyper-parameter values is left to future work.

Since all of our datasets are from the biomedical domain, we found out empirically (see Section \ref{sec:vanilla_vs_pubmed} for the relevant experiment) that it was beneficial to first fine-tune a GPT-2 model on 500k PubMed abstracts, followed by a second round of fine-tuning per dataset, per relation type.

In all cases, we used a pre-trained BERT model (the largest uncased model) as a RE classifier, which we fine-tuned on either the gold or the gold+generated datasets. We used the AdamW optimizer \citep{loshchilov2017fixing}, a sequence length of 128, a batch size of 32 and a learning rate of $2e-5$, We fine-tuned for 5 epochs, keeping the best model with respect to the validation set loss. Also, we used a softmax layer to output predictions and we assigned a relation type to each instance $si$ as follows:
\begin{equation}
    rel\_type(s_i) = 
    \left\{
	\begin{array}{ll}
		argmax(p_c)  & \mbox{if } max(p_c) \geq t \\
		null & \mbox{if } max(p_c) < t
	\end{array}
\right.
\end{equation}

where $c\in L$ and $0 < t < 1$ is a threshold that maximizes the micro-F score on the validation set.

For DARE, in all experiments we train an ensemble of twenty classifiers, where each classifier has been trained on the full gold set and a sub-sample of the generated data. In this way, we manage to alleviate the effect of potential noisy generated instances.

\subsection{Datasets}

To evaluate DARE, we employ three RE datasets from the biomedical domain, their statistics being provided in Table \ref{tbl:datasets}.

The BioCreative V CDR corpus \citep{li2016biocreative} contains chemical-disease relations. The dataset is a binary classification task with one relation type, \textit{chemical induces disease}, and annotations are at the document level, having already been split into train, development and test splits. For simplicity, we followed the work of \citet{papanikolaou2019deep} and considered only intra-sentence relations. We have included the dataset in our GitHub repository to ease replication. In the following, we dub this dataset as \textit{CDR}.

The DDIExtraction 2013 corpus \citep{segura2013semeval} contains MedLine abstracts and DrugBank documents describing drug-drug interactions. The dataset has four relation types and annotations are at the sentence level. The dataset is provided with a train and test split for both MedLine and DrugBank instances. Following previous works, we concatenated the two training sets into one. Also, we randomly sampled 10\% as a development set. In the following this dataset will be referred to as \textit{DDI2013}.

The BioCreative VI-ChemProt corpus \citep{krallinger2017overview} covers chemical-protein interactions, containing five relation types, the vast majority of them being at the sentence level. The dataset comes with a train-development-test split. In the following we will refer to it as \textit{ChemProt}.

\begin{table}
\centering
\begin{tabular}{cccc}
\hline 
GPT-2&Precision&Recall&F1 \\ 
\hline
\hline
Vanilla&\textbf{0.71} &0.69 &0.70\\ 
fine-tuned&0.68&\textbf{0.75}&\textbf{0.73} \\ 
\hline 
\end{tabular}
\caption{\label{tbl:vanilla_vs_pubmed} DARE results on CDR when using a vanilla GPT-2 model or a model that has first been fine-tuned on 500k abstracts from PubMed. In either case the resulting model is then fine-tuned per relation type to generate new examples.}
\end{table}

\subsection{Baselines}

The above datasets suffer both from class imbalance and a limited number of positives. For example the rarest relation type in DDI2013 has only 153 instances in the training set, while the respective one in ChemProt has only 173 data points. Therefore, we consider two suitable baselines for such scenarios, the balanced bagging approach and the class weighting method, both described in Section \ref{sec:related_work}. Both baselines use the base classifier described in Section \ref{sec:setup}. Also, in both cases we consider an ensemble of ten models\footnote{We considered up to 20 models in initial experiments, but there is hardly any improvement after even five models, since the data are repeated.}. Finally, for the class weighting approach we set each class's weight as 
\begin{equation}
    weight_c = \frac{freq_{min}}{freq_c}
\end{equation}
with $min$ being the rarest class.

\subsection{Fine-tuning GPT-2 on In-domain Data }
\label{sec:vanilla_vs_pubmed}
Since all our datasets come from the biomedical domain, we hypothesized that a first round of fine-tuning GPT-2 on in-domain data could be beneficial instead of directly employing the vanilla GPT-2 model. We designed a short experiment using the CDR dataset to test this hypothesis. To clarify, any of the two models (i.e, the vanilla and the one finetuned in in-domain data) would then be fine-tuned per relation type to come up with the final GPT-2 models that would generate the new training examples.

Table \ref{tbl:vanilla_vs_pubmed} illustrates the results of this experiment. As we expect, this first round of fine-tuning proves significantly favourable. We note that when inspecting the generated examples from the vanilla GPT-2 model, generated sentences often contained a peculiar mix of news stories with the compound-disease relations.

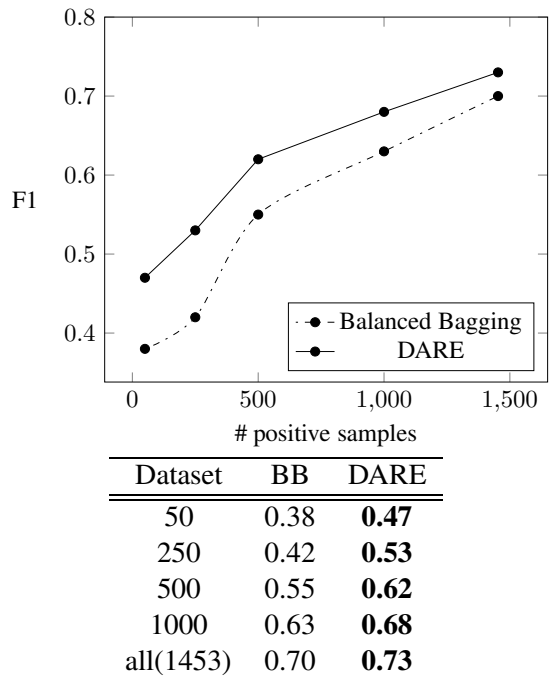
\begin{figure}
   \centering
\begin{tikzpicture}[xscale=0.85,yscale=0.85]
  \begin{axis}[ 
    xlabel={\# positive samples},
    ylabel={F1},
    ymax=0.8,
    xmax=1650,
    legend pos=south east,
    ylabel style={rotate=-90},
  ] 
    	\addplot[                color=black,
                dash pattern=on 1pt off 3pt on 3pt off 3pt,
                mark=*,
                mark options={solid},
                smooth
                ] 
                coordinates {
(50,  0.38)
(250, 0.42)
(500, 0.55)
(1000,    0.63)
(1453,    0.70)
	};
	\addlegendentry{Balanced Bagging}
	
	    	\addplot[color=black, mark=*, mark options={solid},] coordinates {
(50,  0.47)
(250, 0.53)
(500, 0.62)
(1000,    0.68)
(1453,    0.73)
	};
\addlegendentry{DARE}

  \end{axis}
\end{tikzpicture}

\begin{tabular}{ccc}
\hline 
Dataset&BB& DARE\\ 
\hline
\hline
50&0.38&\textbf{0.47}\\
250&0.42&\textbf{0.53}\\
500&0.55&\textbf{0.62}\\
1000&0.63&\textbf{0.68}\\
all(1453)&0.70&\textbf{0.73}\\
\hline 
\end{tabular}
 \caption{DARE vs balanced bagging(BB) for different sizes of positive samples on CDR dataset. Both methods employ ensembles of BERT RE classifiers.}
\label{fig:gold_data_size_study}
\end{figure}

\subsection{DARE on Imbalanced Datasets}
\label{sec:size}
In this experiment, we wanted to evaluate the effect of our method when dealing with great imbalance, i.e., datasets with very few positive samples. To that end, we considered the CDR dataset and sampled different numbers of positive examples from the dataset (50, 250, 500, 1000 and all positives) and combined them with all the negative instances. The resulting five datasets were used to train either a balanced bagging ensemble or DARE. 

In Figure \ref{fig:gold_data_size_study}, we show the results, averaging across five different runs. In all cases, our approach has a steady, significant advantage over the balanced bagging baseline, their difference reaching up to 11 F1 score points when only few positives ($\leq 250$) are available. As we add more samples, the differences start to smooth out as expected. These results clearly illustrate that DARE can boost the predictive power of a classifier when dealing with few positive samples, by cheaply generating training data of arbitrary sizes.

\begin{figure}
   \centering
\begin{tikzpicture}[xscale=0.85,yscale=0.85]
  \begin{axis}[ 
    xlabel=$ratio$,
    ylabel={F1},
    ylabel style={rotate=-90},
  ] 
    	\addplot[                color=black,
                dash pattern=on 1pt off 3pt on 3pt off 3pt,
                mark=*,
                mark options={solid},
                smooth
                ]  coordinates {
(0.5, 0.66)
(1,  0.68)
(2,  0.64)
(4,  0.62)
	};
	\addlegendentry{DARE}
	    	\addplot[color=black, mark=*, mark options={solid},] coordinates {
(0.5, 0.63)
(1,  0.63)
(2,  0.63)
(4,  0.63)
	};
	\addlegendentry{Balanced Bagging}
  \end{axis}
\end{tikzpicture}
 \caption{DARE performance for different generated dataset sizes in each base classifier. For each relation type we add \textit{ratio}$*|D_c|$ examples.}
\label{fig:study}
\end{figure}
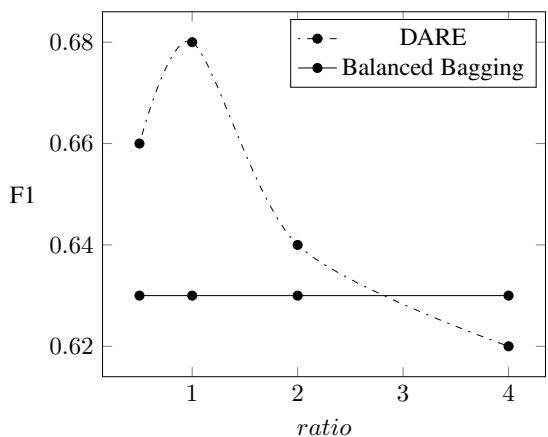

\subsection{Effect of Generated Data Size}
\label{sec:ratio_study}

Our next experiment focuses in studying the effect of different sizes of generated data on DARE's performance.

As explained, our method relies on fine-tuning GPT-2 to generate examples for each relation type that will, ideally, come from the same distribution as the ones from the gold training data. Nevertheless, we should expect that this procedure will not be perfect, generating also noisy samples. As mentioned previously, we try to alleviate this effect by training an ensemble of classifiers, each trained on the whole gold and a part of the generated dataset. 

An important question that arises therefore, is how to determine the optimal ratio of generated examples to include in each classifier. If too few, the improvements will be insignificant, if too many we risk to have the model being influenced by the noise. 

In order to gain empirical insight into the above question we design a short experiment using the CDR dataset, for different sizes of generated data. As gold set, we consider a random subset of 1,000 positive examples and all negatives, to make more prominent the effect of class imbalance. 

In Figure \ref{fig:study} we show the results for five different generated data sizes. Interestingly, adding more data does not necessarily boost classifier performance, since the noisy patterns in the generated data seem to influence more the classifier than those in the gold data. In the following, we choose a $ratio=1$, adding for each relation type a number of generated instances equal to the number of gold instances. It should be noted that we are not limited in the total generated data that we will use since we can fine-tune an arbitrary number of classifiers on combinations of the gold data and subsets of the generated data.

\begin{table*}
\centering
\begin{tabular}{ccccc}
\hline 
Dataset&Configuration&Precision&Recall&F1 \\ 
\hline
\hline
CDR&SOTA \citep{papanikolaou2019deep}&0.61&0.80&0.70 \\ 
&BERT+class weighting&0.66&0.74&0.69 \\ 
&BERT+balanced bagging&0.61&0.79&0.70 \\ 
&DARE&0.68&0.75&\textbf{0.73} \\ 
\hline
ChemProt&SOTA \citep{peng2018extracting}&0.72&0.58&0.65 \\ 
&BERT+class weighting&0.75&0.67&0.70 \\ 
&BERT+balanced bagging&0.69&\textbf{0.71}&0.70 \\ 
&BERT+DARE&\textbf{0.79}&0.68&\textbf{0.73} \\ 
\hline
DDI2013&SOTA  \citep{sun2019drug}& 0.77& 0.74& 0.75\\
&BERT+class weighting&0.81&0.71&0.76\\ 
&BERT+balanced bagging&0.74&0.72&0.73\\ 
&BERT+DARE&0.82&0.74&\textbf{0.78}\\ 
\hline 
\end{tabular}
\caption{\label{tbl:comparison} Comparison of DARE vs the previous SOTA and two baselines suited for imbalanced datasets. Only statistically significant results to the second best model are marked in bold. Statistical significance is determined with a McNemar p-test at 0.05 significance level.}
\end{table*}

\subsection{DARE against the SOTA and Baselines}

Taking into account the previous observations, we proceed to compare DARE against the SOTA and the two previously described baselines. Table \ref{tbl:comparison} describes the results. For the multi-class datasets we report the micro-F score in order to make our results comparable with previous works. Also, in the Supplementary Material 
we report the per class results for DARE against the SOTA and the class weighting baseline, for the two multi-class datasets in order to ease comparison with past or future works.

Comparing DARE against the SOTA, we observe a steady advantage of our method across all datasets, ranging from 3 to 8 F1 points. These results are somehow expected, since we employ BERT-large as our base classifier which has proven substantially better than Convolutional (CNN) or Recurrent neural networks (RNN) across a variety of tasks \citep{devlin2018bert}. In CDR, \citet{papanikolaou2019deep} have used BioBERT\citep{lee2019biobert} which is a BERT base (cased) model pre-trained on PubMed, while we use BERT large (uncased), in ChemProt, \citet{peng2018extracting} use ensembles of SVM, CNN and RNN models while in DDI2013 \citet{sun2019drug} have used hybrid CNN-RNN models.

When observing results for the baselines, we notice that they perform roughly on par. DARE is better from 2 to 5 F1 points against the baselines, an improvement that is smaller than that against the SOTA, but still statistically significant in all cases.

Overall, and in accordance with the results from the experiment in Section \ref{sec:size}, we observe that DARE manages to leverage the GPT-2 automatically generated data, to steadily improve upon the SOTA and two competitive baselines.

\section{Conclusions}
\label{sec:conclusions}
We have presented DARE, a novel method to augment training data in Relation Extraction. Given a gold RE dataset, our approach proceeds by fine-tuning a pre-trained GPT-2 model per relation type and then uses the fine-tuned models to generate new training data. We sample subsets of the synthetic data with the gold dataset to fine-tune an ensemble of RE classifiers that are based on BERT. Through a series of experiments we show empirically that our method is particularly suited to deal with class imbalance or limited data settings, recording improvements up to 11 F1 score points over two strong baselines. We also report new SOTA performance on three biomedical RE benchmarks.

Our work can be extended with minor improvements on other Natural Language Understanding tasks, a direction that we would like to address in future work.

\bibliographystyle{acl_natbib}
\bibliography{main}

\newpage
\appendix

\section{Supplemental Material}
\label{sec:supplemental}
In this section we present additionally the results per class for ChemProt and DDI2013, for DARE against the class weighting baseline and the SOTA.
\begin{table}[h]
\centering
\begin{tabular}{cccc}
\hline 
relation type&SOTA&Class Weighting&DARE \\
\hline
\hline
CPR-3&-  &    0.66  &    \textbf{0.70}\\
CPR-4&-  &    0.75&      \textbf{0.79}\\
CPR-5&-  &    0.73    &  \textbf{0.81}\\
CPR-6& - &     0.69 &     \textbf{0.73}\\
CPR-9&-  &    0.57&     \textbf{0.59}\\
\hline 
\end{tabular}
\caption{\label{tbl:chemprot} ChemProt results per relation type for DARE vs SOTA and best baseline in terms of F1.}
\end{table}

\begin{table}[h]
\centering
\begin{tabular}{cccc}
\hline 
relation type&SOTA&Class Weight&DARE \\
\hline
\hline
advise&\textbf{0.81} &  \textbf{0.81}    &     0.80\\
effect&0.73 &  0.76     &    \textbf{0.78}\\
int&\textbf{0.59}   &   0.52   &  0.58\\
mechanism& 0.78 &     0.78   &  \textbf{0.80}\\
\hline 
\end{tabular}
\caption{\label{tbl:DDI} DDI2013 results per relation type for DARE vs state-of-the-art and best baseline in terms of F1.}
\end{table}

\end{document}